\documentclass{ieeeaccess}
\usepackage{cite}
\usepackage{amsmath,amsfonts}
\usepackage{algorithmic}
\usepackage{graphicx}
\usepackage{textcomp}
\usepackage{hyperref}

\usepackage{bm}
\makeatletter
\AtBeginDocument{\DeclareMathVersion{bold}
\SetSymbolFont{operators}{bold}{T1}{times}{b}{n}
\SetSymbolFont{NewLetters}{bold}{T1}{times}{b}{it}
\SetMathAlphabet{\mathrm}{bold}{T1}{times}{b}{n}
\SetMathAlphabet{\mathit}{bold}{T1}{times}{b}{it}
\SetMathAlphabet{\mathbf}{bold}{T1}{times}{b}{n}
\SetMathAlphabet{\mathtt}{bold}{OT1}{pcr}{b}{n}
\SetSymbolFont{symbols}{bold}{OMS}{cmsy}{b}{n}
\renewcommand\boldmath{\@nomath\boldmath\mathversion{bold}}}
\makeatother

\def\BibTeX{{\rm B\kern-.05em{\sc i\kern-.025em b}\kern-.08em
    T\kern-.1667em\lower.7ex\hbox{E}\kern-.125emX}}

\begin{document}
\history{Date of publication July 24, 2024, date of current version July 24, 2024.}
\doi{10.1109/ACCESS.2024.0429000}

\title{Advanced Wildfire Prediction in Morocco: Developing a Deep Learning Dataset from Multisource Observations}
\author{\uppercase{Ayoub Jadouli}\authorrefmark{1}and \uppercase{Chaker El Amrani}\authorrefmark{1} \IEEEmembership{Member, IEEE}}
\address[1]{Computer Science and Smart Systems, Faculty of Sciences and Technology, Abdelmalek Essaâdi University, Tangier, Morocco (e-mail: ajadouli@uae.ac.ma)}

\markboth
{Author \headeretal: Preparation of Papers for IEEE TRANSACTIONS and JOURNALS}
{Author \headeretal: Preparation of Papers for IEEE TRANSACTIONS and JOURNALS}

\corresp{Corresponding author: Ayoub Jadouli (e-mail: ajadouli@uae.ac.ma).}

\begin{abstract}
Wildfires pose significant threats to ecosystems, economies, and communities worldwide, necessitating advanced predictive methods for effective mitigation. This study introduces a novel and comprehensive dataset specifically designed for wildfire prediction in Morocco, addressing its unique geographical and climatic challenges. By integrating satellite observations and ground station data, we compile essential environmental indicators—such as vegetation health (NDVI), population density, soil moisture levels, and meteorological data—aimed at predicting next-day wildfire occurrences with high accuracy. Our methodology incorporates state-of-the-art machine learning and deep learning algorithms, demonstrating superior performance in capturing wildfire dynamics compared to traditional models. Preliminary results show that models using this dataset achieve an accuracy of up to 90\%, significantly improving prediction capabilities. The public availability of this dataset fosters scientific collaboration, aiming to refine predictive models and develop innovative wildfire management strategies. Our work not only advances the technical field of dataset creation but also emphasizes the necessity for localized research in underrepresented regions, providing a scalable model for other areas facing similar environmental challenges.
\end{abstract}

\begin{keywords}
Deep learning, machine learning, wildfire prediction, satellite observations, ground stations, Morocco, NDVI, soil moisture.
\end{keywords}

\titlepgskip=-21pt

\maketitle

\section{Introduction}
In the realm of environmental science and disaster management, the development of predictive models for wildfires has become increasingly critical due to their significant impact on ecosystems, economies, and communities globally \cite{Chergui2018, FAO2017}. In regions like Morocco, where data availability is often constrained, the need for specialized datasets tailored to local geographical and climatic conditions is paramount \cite{Ganteaume2013}. Morocco's diverse ecosystems, ranging from Mediterranean forests to arid deserts, present unique challenges for wildfire prediction, underscoring the importance of localized data collection and modeling efforts \cite{Moritz2014}.

Existing wildfire prediction methods have various limitations. Traditional models often fail to capture the complex interactions between different environmental factors influencing wildfire occurrences \cite{Padilla2011}. Machine learning (ML) and deep learning (DL) approaches have shown promise in enhancing predictive accuracy by leveraging large-scale datasets and sophisticated algorithms \cite{Pettinari2016}. However, these models require high-quality, relevant datasets that encapsulate the myriad factors influencing wildfire dynamics \cite{Rodrigues2020}.

To address these challenges, we have curated a novel dataset specifically for Morocco, integrating detailed environmental and meteorological data essential for accurate wildfire prediction. This dataset includes vegetation health (NDVI), population density, soil moisture levels, and various meteorological variables, providing a comprehensive basis for predictive modeling \cite{Scott2013, Smith2011}. Additionally, the dataset has been structured to be columnar, easy to use, and compress multiple dimensions into a single format, facilitating its use in ML applications.

One of the significant barriers to advancing wildfire prediction in regions like Morocco is the lack of readily available, well-prepared datasets that researchers can easily utilize. By creating a dataset that is both comprehensive and user-friendly, we aim to fill this gap and support the scientific community in developing more effective predictive models.

The primary contributions of this study are threefold:
\begin{enumerate}
    \item We present a novel, publicly available dataset tailored to Morocco's unique environmental conditions, facilitating research in underrepresented regions.
    \item We demonstrate the dataset's effectiveness in capturing wildfire dynamics through extensive testing with advanced ML and DL models.
    \item We provide a scalable model for other regions facing similar environmental challenges, promoting scientific collaboration and innovation in wildfire management strategies.
\end{enumerate}

In the following sections, we detail the methodology behind the dataset's creation, its potential applications for wildfire prediction in Morocco, and its role in advancing environmental modeling and disaster preparedness. By bridging the data gap and enabling the benchmarking of predictive models, our work lays the foundation for more informed and effective wildfire management strategies.

\section{Related Work}

The integration of machine learning (ML) and deep learning (DL) techniques in wildfire prediction has significantly advanced the field, providing more accurate and timely predictions. This section reviews relevant works, critically analyzes their limitations, and positions our contribution in the context of existing research.

\subsection{Existing Datasets for Wildfire Prediction}

\subsubsection{Global and Regional Datasets}
Numerous datasets have been developed to support wildfire prediction across different regions. Huot et al. \cite{Huot2022} introduced the "Next Day Wildfire Spread" dataset, combining a decade of U.S. remote-sensing data tailored for ML applications. While this dataset demonstrates the efficacy of neural networks, its reliance on U.S.-centric data limits its applicability to other regions.

Jain et al. \cite{Jain2020} reviewed various ML applications in wildfire science, emphasizing the importance of integrating large-scale datasets with advanced computational techniques. However, they did not address the challenges of data heterogeneity and quality, which are crucial for accurate predictions.

Khan et al. \cite{Khan2022} proposed the "DeepFire" dataset, containing UAV-based forest imagery with and without fire. This dataset, consisting of 1900 colored images, assists in fire detection using supervised ML classifiers and a VGG19-based transfer learning approach. However, the dataset lacks geographical specificity, which is critical for localized predictions.

Kondylatos et al. \cite{Kondylatos2022} developed a dataset for wildfire danger prediction using DL models. Their work demonstrates the potential of such datasets in improving prediction accuracy but does not fully address the complexities of regional environmental dynamics.

\subsubsection{Challenges in the Moroccan Context}
Despite the availability of global datasets, there is a scarcity of region-specific datasets for wildfire prediction in Morocco. The unique geographical and climatic conditions in Morocco necessitate tailored datasets that can accurately capture local environmental dynamics. Previous studies, such as those by Mharzi-Alaoui et al. \cite{MharziAlaoui2022}, have highlighted the challenges and opportunities in applying ML to wildfire management in Morocco, demonstrating the need for localized data collection and modeling efforts.

\subsection{Comparison with Other Contributions}

Our study addresses the gap in localized datasets by introducing a comprehensive dataset specifically curated for Morocco. Key differences and advantages of our dataset compared to existing contributions include:

\begin{itemize}
\item \textbf{Geographical Specificity:} Unlike global datasets, our dataset is tailored to the unique environmental conditions of Morocco, capturing critical factors such as vegetation health (NDVI), population density, soil moisture levels, and meteorological data.
\item \textbf{Data Integration:} We integrate multiple data sources, including satellite observations and ground station data, to create a multidimensional dataset that supports comprehensive analysis and prediction.
\item \textbf{Columnar Data:} Our dataset is based on  columnar data refined from the ground station and Level 3 (L3) satellite data, which is less performance-hungry compared to image-based datasets used in other studies, such as those by Khan et al. \cite{Khan2022}. The processing of columnar data is more efficient and feasible for deployment in resource-constrained environments like Morocco.
\item \textbf{Public Availability:} Our dataset is publicly available on Kaggle \cite{JADOULI2024}, encouraging collaboration and further research in wildfire prediction and management.
\item \textbf{Feature Augmentation:} We employ advanced feature selection and augmentation techniques to enhance the dataset's predictive capabilities, ensuring it captures the complex dynamics of wildfire occurrences.
\end{itemize}

\subsection{Machine Learning and Deep Learning Applications}

The application of ML and DL models to our dataset demonstrates significant improvements in predictive accuracy. Models such as LightGBM \cite{Ke2017} and XGBoost \cite{Chen2016} have shown robust performance, while DL architectures have further enhanced prediction capabilities by capturing complex patterns in the data.

Our study also employs permutation feature importance analysis to identify the most influential features in predicting wildfire occurrences. This analysis reveals the critical role of geographical and environmental factors, such as latitude and NDVI, in determining wildfire risk.

Additionally, Jadouli and El Amrani \cite{Jadouli2022} presented a study on the detection of human activities in wildlands to prevent wildfires. Using deep learning on remote sensing images, their work demonstrated the effectiveness of convolutional neural networks (CNN) in classifying images to detect human interactions with wildlands, which are primary causes of wildfires. Their approaches include object detection, scene classification, and land class differentiation, achieving significant accuracy in detecting human activities that can lead to wildfires.

The Authors have also shown the efficacy of combining multisource spatio-temporal data with deep learning and ensemble models for improving wildfire forecasting accuracy. The work by Jadouli and El Amrani \cite{Jadouli2024Ensemble} leverages satellite data and transfer learning to enhance wildfire prediction, demonstrating significant improvements over traditional methods.

By addressing the specific challenges in the Moroccan context, our dataset and methodologies offer significant improvements over traditional methods, advancing the field of wildfire prediction and management.

\section{Methodology Overview}

This section provides a detailed explanation of the methodologies employed to predict wildfire occurrences, integrating various datasets and analytical techniques. The comprehensive approach ensures robust data collection, pre-processing, feature selection, and machine learning model implementation. The code used is available for review and use on Kaggle and GitHub \cite{JADOULI2024, JADOULI2024Jupyter, Github2024}.

\subsection{Data Collection and Integration}

Predictive analysis of wildfire occurrences requires a multidimensional approach to data collection and integration. This section details the systematic methodologies employed to gather, preprocess, and integrate diverse datasets, establishing a robust foundation for subsequent analysis \cite{BigQuery2024}.

\subsubsection{Data Sources and Preprocessing}

\paragraph{Weather Data Collection and Preprocessing}

Weather conditions significantly influence wildfire dynamics, making the integration of comprehensive meteorological data essential. Our primary source, the Global Surface Summary of the Day (GSOD) dataset provided by NOAA's National Centers for Environmental Information, offers extensive daily meteorological observations from a global network of weather stations \cite{Smith2011}.

\textbf{Data Collection Process:}
Utilizing BigQuery, we extracted weather data pertinent to Moroccan stations from the GSOD dataset for the period 2010 to 2023. This process entailed querying the database annually and amalgamating the resulting datasets into a singular DataFrame, thereby facilitating a streamlined analysis \cite{BigQuery2024}.

\textbf{Preprocessing Steps:}
The weather data underwent meticulous preprocessing to extract relevant features (average temperature, maximum temperature, precipitation, wind speed), standardize column names, and address missing data through interpolation, ensuring the dataset's integrity and usability \cite{DataPrep2024}.

\paragraph{Incorporation of NASA FIRMS Wildfire Data}

The Fire Information for Resource Management System (FIRMS) by NASA provides near-real-time active fire data, crucial for our analysis. Automated procedures were implemented to download fire event records from 2010 to 2023 using a Python script, utilizing the Requests library for efficient data retrieval \cite{NASA2024}.

\textbf{Data Loading and Processing:}
FIRMS data, detailing latitude, longitude, brightness temperatures, and other critical attributes, was systematically loaded into pandas DataFrames. Annual datasets for both MODIS and VIIRS instruments were concatenated, creating comprehensive records for analysis \cite{Pandas2024}.

\textbf{Geospatial Analysis:}
Geospatial analysis, facilitated by the OSMnx library, enabled the integration of FIRMS data with geographical information, significantly enriching our dataset with context on potential human impacts \cite{Geospatial2024}.

\subsubsection{Integration with Environmental and Temporal Data}

\paragraph{Human Population and Temporal Data Integration}

The "UN WPP-Adjusted Population Density, v4.11" dataset provided detailed population distribution patterns. Preprocessing involved extracting population density information for Morocco, enhancing our dataset with high-resolution spatial data on human settlements \cite{UN2024}. Public holidays and day-of-the-week data were also integrated, recognizing their impact on human activities relevant to wildfire risks. A dedicated dataset was constructed, translating various date formats into a consistent datetime format for accurate temporal analysis.

\paragraph{Integration of Vegetation and Soil Moisture Data}

\textbf{AMSR2/GCOM-W1 Surface Soil Moisture Data:}
To enhance our understanding of environmental conditions affecting wildfire occurrences, we integrated the "AMSR2/GCOM-W1 surface soil moisture (LPRM) L3 1 day 25 km x 25 km descending V001" dataset, spanning from 2012 to 2024. This dataset, provided by the NASA Goddard Earth Sciences Data and Information Services Center (GES DISC), includes crucial measurements of surface soil moisture content, offering a detailed record essential for our analysis \cite{NASAAMSR2024}.

\textbf{NDVI Data for Vegetation Health:}
Additionally, the study leverages NDVI data from 2010 to 2024, obtained from the "ndvi3g\_geo\_v1\_2\_\_*.nc4" dataset. NDVI is a widely recognized index for assessing vegetation health and productivity, derived from NASA Earth's Advanced Very High Resolution Radiometer (AVHRR) data \cite{NASANDVI2024}.

\subsection{Feature Selection and Augmentation}

The accuracy of predictive modeling in assessing wildfire risks hinges on the careful selection of relevant features and the augmentation of data to reflect the complex dynamics of wildfire occurrences. This section outlines the rationale behind the selection of predictive features from our integrated dataset and describes the strategies employed to enrich the data, thereby augmenting the predictive capabilities of our models.

\subsubsection{Feature Selection}

The selection of features for predicting wildfire occurrences was guided by an extensive review of literature and empirical analysis, aiming to identify variables with significant predictive power. The features chosen encompass a wide range of environmental, meteorological, and human-related factors, reflecting the multifaceted nature of wildfire dynamics.

\textbf{Meteorological Features:}
Temperature, precipitation, wind speed, and humidity were selected due to their direct impact on wildfire ignition and spread. Historical weather data, specifically the 15-day lagged values for each variable, were included to capture the immediate historical context leading up to wildfire events \cite{Smith2011, DataPrep2024}.

\textbf{Vegetation and Soil Moisture:}
NDVI and soil moisture levels serve as critical indicators of vegetation health and fuel moisture, respectively. These features are essential for assessing the availability and combustibility of wildfire fuels \cite{NASANDVI2024}.

\textbf{Human-related Features:}
Population density and proximity to urban areas were integrated to evaluate human influence on wildfire risks. Data on public holidays and weekends were included to consider variations in human activity that might affect wildfire occurrences \cite{UN2024}.

\textbf{Temporal Features:}
The inclusion of day-of-the-week and month variables allows the model to account for seasonal and weekly patterns in wildfire occurrence, acknowledging the influence of climatic conditions and human activities on fire dynamics over time \cite{DataPrep2024}.

\subsubsection{Augmented Data}

Augmenting the dataset was critical for improving the granularity and coverage of our wildfire data, enabling a more nuanced understanding of the spatial and temporal patterns of wildfire risks.

\textbf{Enriched Wildfire Data:}
For each entry in the wildfire\_df\_simplified DataFrame, we generated 300 nearby points within a 300-meter radius to simulate potential wildfire occurrences around the original fire points. This approach aimed to capture the spatial continuity of environmental conditions affecting wildfire risks, enhancing the dataset with a denser representation of the wildfire landscape.

\textbf{Non-Fire Data Generation:}
To provide a contrastive dataset for model training, non-fire data points were generated and distributed across Morocco, avoiding immediate vicinities of known wildfire occurrences. This dataset was instrumental in distinguishing areas of high wildfire risk from less susceptible ones, based on predictive variables.

\subsection{Methodological Considerations}

This section outlines the sophisticated methodological approaches adopted to process, expand, and integrate various datasets into our wildfire prediction model. These methodologies not only bolster the robustness of our predictive analysis but also address potential challenges arising from the complex nature of the data involved \cite{DataPrep2024}.

\subsubsection{Data Expansion and Integration Techniques}

To capture the intricate dynamics influencing wildfire risks, we employed several advanced data expansion and integration techniques. These methods are designed to enrich our dataset, providing a comprehensive view of the factors that influence wildfire occurrences.

\textbf{Weather Data Expansion:}
A key strategy was the creation of lagged features for each meteorological variable, extending up to 15 days prior to any given event. This historical context is crucial for understanding the buildup of conditions conducive to wildfires. We also computed aggregate statistics (weekly, monthly, quarterly, yearly means) for meteorological variables to capture long-term climatic trends and their impact on wildfire risks.

\textbf{Integration with Other Datasets:}
The weather dataset was enhanced through integration with additional data sources, including population density, NDVI, and soil moisture levels. This multidimensional approach allows for a holistic analysis of wildfire risks by incorporating human, environmental, and meteorological factors.

\textbf{Temporal Resolution Enhancement:}
Beyond immediate historical data, our model incorporates weather conditions from the previous year that align with the same week or month of a current event. This approach accounts for seasonal climatic variability, further refining the model’s predictive capabilities.

\subsubsection{Proximity to Maritime Boundaries}

Understanding the geographical context is vital for comprehensive wildfire risk assessment. Our methodology includes calculating the shortest distance from each observation point to the nearest maritime boundary, employing geospatial analysis techniques facilitated by the Geopandas library.

\textbf{Methodological Approach:}
We transformed our observational dataset into a GeoDataFrame and utilized a comprehensive GeoDataFrame of global maritime boundaries for distance calculations. This analysis helps in understanding the influence of proximity to water bodies on wildfire occurrences. Distances were calculated using the World Mercator projection system, which balances ease of use with the precision required for our analysis, despite its limitations in area distortion at extreme latitudes.

\subsubsection{Addressing Data Heterogeneity and Scale}

The integration of diverse datasets presents challenges, particularly in terms of varying data scales and formats. We addressed these challenges through meticulous pre-processing, including standardization of spatial resolutions and alignment of temporal scales across datasets. This ensured seamless integration and analysis of data from disparate sources.

\textbf{Data Standardization:}
Conversion of satellite-derived data into a structured format suitable for analysis involved aligning spatial resolutions and standardizing data formats, ensuring consistency across the integrated dataset.

\textbf{Temporal Alignment:}
Temporal data from various sources were aligned by aggregating or interpolating data to match the temporal resolution of key datasets, facilitating a coherent analysis over time.

\subsubsection{Ethical Considerations in Data Usage}

In conducting this study, we adhered to ethical guidelines concerning data privacy and usage, especially when dealing with human-related data such as population density and activity patterns. All data used were sourced from public datasets or obtained with appropriate permissions, ensuring compliance with data protection regulations.

\subsection{Final Dataset Structure and Balance}

The comprehensive approach undertaken in this study has culminated in the creation of a rich, multidimensional dataset designed to predict wildfire occurrences with high accuracy. This section outlines the final structure of the dataset, including a detailed list of columns that capture a wide range of environmental, meteorological, and human-related factors. Additionally, we address the balance achieved in the dataset regarding the binary outcome variable \textit{is\_fire}, indicating the presence or absence of a wildfire.

\textbf{Dataset Columns:}
The final dataset encompasses the following columns, each providing critical insights into the conditions surrounding wildfire occurrences:
\begin{itemize}
    \item \textbf{Temporal Features:} \textit{acq\_date}, \textit{is\_holiday}, \textit{day\_of\_week}, \textit{day\_of\_year}, \textit{is\_weekend}
    \item \textbf{Spatial Features:} \textit{latitude}, \textit{longitude}, \textit{station\_lat}, \textit{station\_lon}, \textit{sea\_distance}
    \item \textbf{Environmental Indicators:} \textit{NDVI}, \textit{SoilMoisture}
    \item \textbf{Meteorological Data (Lagged Variables):} Includes lagged variables for \textit{average\_temperature}, \textit{maximum\_temperature}, \textit{minimum\_temperature}, \textit{precipitation}, \textit{snow\_depth}, \textit{wind\_speed}, \textit{maximum\_sustained\_wind\_speed}, \textit{wind\_gust}, \textit{dew\_point}, \textit{fog}, \textit{thunder} for lags 1 to 15 days.
    \item \textbf{Additional Meteorological Measures:} Weekly, monthly, quarterly, and yearly means for key weather variables, alongside lagged weather data from the previous 1 to 3 years, to capture both immediate and historical climatic influences.
    \item \textbf{Outcome Variable:} \textit{is\_fire}, indicating whether a wildfire occurred (1) or not (0).
\end{itemize}

\textbf{Dataset Shape:}
The dataset is structured to facilitate advanced analytics, containing numerous columns that reflect the diverse range of factors influencing wildfire risks. This structure supports the application of machine learning algorithms capable of processing complex interactions among variables to predict wildfire occurrences accurately.

\textbf{Balance in the Dataset:}
To ensure the effectiveness of predictive modeling, particularly for binary classification tasks, it's crucial to address potential imbalances in the outcome variable. In our dataset, the \textit{is\_fire} variable, indicating the presence or absence of wildfires, has been balanced meticulously to prevent model bias towards the more frequent class. The dataset contains:
\begin{itemize}
    \item \textit{is\_fire = 0}: 467,293 instances
    \item \textit{is\_fire = 1}: 467,293 instances
\end{itemize}
This balanced distribution ensures that the predictive models developed from this dataset are not biased towards predicting one outcome over the other, enhancing the reliability and accuracy of wildfire occurrence predictions.

\subsection{Machine Learning Methods}

\subsubsection{Overview}

To benchmark the performance of various machine learning models in predicting wildfire occurrences, we utilized several algorithms with GPU support. This approach leverages the computational power of GPUs to accelerate the training and inference processes, making it feasible to handle large datasets efficiently \cite{JADOULI2024Jupyter}.

\subsubsection{Libraries and GPU Utilization}

The primary library used for this benchmarking is cuML, a suite of GPU-accelerated machine learning algorithms provided by NVIDIA's RAPIDS project \cite{RAPIDS2018}. Additionally, we employed XGBoost \cite{Chen2016} and LightGBM \cite{Ke2017}, both of which support GPU acceleration for enhanced performance. These libraries allow for faster training times and improved efficiency, especially when dealing with large-scale data.

\subsubsection{Logistic Regression, SVM, K-Nearest Neighbors, and Random Forest}

We implemented several classic machine learning models using cuML, which provides GPU-accelerated versions of these algorithms. The models used include:
\begin{itemize}
    \item \textbf{Logistic Regression:} A linear model used for binary classification that predicts the probability of a binary outcome.
    \item \textbf{Support Vector Machine (SVM):} A model that finds the optimal hyperplane to separate classes in the feature space.
    \item \textbf{K-Nearest Neighbors (KNN):} A non-parametric method used for classification by comparing a point with its nearest neighbors.
    \item \textbf{Random Forest:} An ensemble learning method that constructs multiple decision trees for classification tasks.
\end{itemize}
These models were trained and validated using the available data, with performance metrics such as accuracy, Area Under the Curve (AUC), precision, and recall being recorded to assess their effectiveness.

\subsubsection{XGBoost}

XGBoost is an optimized gradient boosting library designed to be highly efficient and flexible. It includes support for GPU acceleration, which significantly speeds up the training process \cite{Chen2016}. We used XGBoost to build a robust model by leveraging its ability to handle missing data, support regularization, and efficiently manage overfitting. The model was trained using specific hyperparameters optimized for performance, and its effectiveness was evaluated using accuracy, AUC, precision, and recall metrics.

\subsubsection{LightGBM}

LightGBM is a highly efficient gradient boosting framework that uses tree-based learning algorithms. It supports GPU acceleration, making it suitable for training on large datasets quickly \cite{Ke2017}. LightGBM's key advantages include its ability to handle large-scale data and its speed in training and prediction. The model was trained and evaluated similarly to the other models, with performance metrics being recorded to compare its effectiveness.

\subsubsection{Deep Learning Model Configuration and Training}

To systematically identify the optimal deep learning model for wildfire prediction, we implemented a comprehensive grid search across various hyperparameters. The code uses the Keras Sequential API to create and evaluate multiple model configurations. Specifically, we explored different combinations of the number of layers (ranging from 3 to 6), units per layer (16, 32, 64, 128, 256, 300), dropout rates (0.6, 0.5, 0.4, 0.3, 0.0), and activation functions (ReLU, GELU, Softplus, LeakyReLU). Each configuration was generated using the product function from itertools and shuffled to ensure randomness in training order. For each configuration, a model was defined with the specified number of layers, units, dropout rate, and activation function. LeakyReLU required specific handling within the loop. Models were compiled with the Adam optimizer and binary cross-entropy loss, and evaluated using accuracy, AUC-PR, precision, and recall metrics. A ModelCheckpoint callback saved the best model based on validation accuracy during training. Each model configuration was trained for 20 epochs with a batch size of 512. After training, the best validation metrics were recorded and stored in a DataFrame, which was continuously updated and saved to a CSV file for further analysis. This methodology enabled a thorough exploration of the hyperparameter space to determine the most effective model architecture.

\begin{figure}[h]
    \centering
    \includegraphics[width=0.2\textwidth]{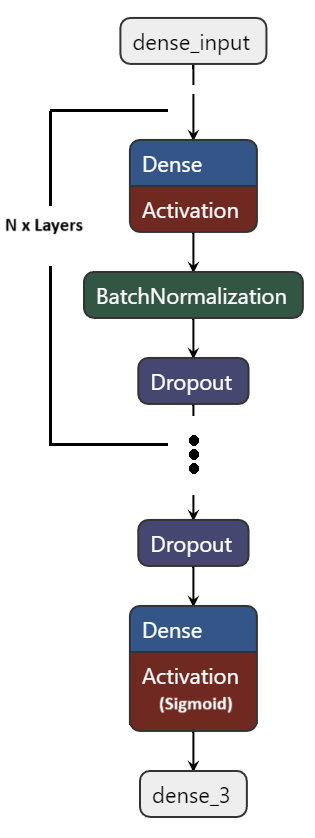}
    \caption{Grid Search Implementation Results Architecture for Deep Learning Model Configurations}
    \label{fig:grid_search}
\end{figure}

\subsection{Permutation Feature Importance Analysis}

To identify the most influential features in predicting wildfire occurrences, we employed the Permutation Feature Importance (PFI) technique. This method involves the following steps:
\begin{itemize}
    \item \textbf{Model Setup:} A feedforward neural network was trained with layers designed to capture nonlinear relationships within the data.
    \item \textbf{Feature Shuffling:} Each feature in the test set was shuffled individually, and the decrease in the model's accuracy was measured, indicating the importance of that specific feature.
    \item \textbf{Importance Calculation:} The difference in accuracy with and without the shuffling of each feature was calculated to determine its impact on model performance.
\end{itemize}
This analysis provides insights into which features significantly affect the model's prediction accuracy, guiding further data collection and feature engineering efforts.

\section{Results}

\subsection{Model Performance}

This section presents the performance of various machine learning models in predicting wildfire occurrences. We evaluate the models using metrics such as accuracy, Area Under the Curve (AUC), average precision, and average recall. The results of validation are detailed in the following tables and discussed subsequently.

\begin{table}[h!]
\centering
\caption{Performance Metrics for Traditional Machine Learning Models}
\begin{tabular}{|c|c|c|c|c|}
\hline
Model & Accuracy & AUC & Precision & Recall \\
\hline
Logistic Regression & 0.5039 & 0.7543 & 0.6091 & 0.6227 \\
SVM & 0.7871 & 0.9364 & 0.8339 & 0.6907 \\
K-Nearest Neighbors & 0.6262 & 0.6771 & 0.8142 & 0.3550 \\
Random Forest & 0.6678 & 0.9350 & 0.7419 & 0.8978 \\
XGBoost & 0.6674 & 0.9061 & 0.6905 & 0.8963 \\
LightGBM & 0.9072 & 0.9501 & 0.6994 & 0.9417 \\
\hline
\end{tabular}
\label{tab:traditional_models}
\end{table}

Table \ref{tab:traditional_models} summarizes the performance of traditional machine learning models. These models include Logistic Regression, SVM, K-Nearest Neighbors, Random Forest, XGBoost, and LightGBM. Each model's performance is assessed based on accuracy, AUC, average precision, and average recall.

\begin{table}[h!]
\centering
\caption{Top Deep Learning Models by Accuracy}
\begin{tabular}{|c|c|c|c|c|}
\hline
Model Configuration & Accuracy & AUC-PR & Precision & Recall \\
\hline
3L-256U-0.6D-relu & 0.8994 & 0.8559 & 0.8895 & 0.9122 \\
5L-128U-0.5D-relu & 0.8954 & 0.8910 & 0.8714 & 0.9277 \\
4L-128U-0.5D-leakyrelu & 0.8938 & 0.8617 & 0.8787 & 0.9138 \\
4L-256U-0.6D-leakyrelu & 0.8935 & 0.8709 & 0.8799 & 0.9114 \\
4L-64U-0.3D-relu & 0.8934 & 0.8573 & 0.8762 & 0.9164 \\
4L-256U-0.4D-relu & 0.8931 & 0.9063 & 0.8782 & 0.9129 \\
4L-300U-0.5D-leakyrelu & 0.8924 & 0.9068 & 0.8728 & 0.9187 \\
5L-256U-0.6D-relu & 0.8922 & 0.8632 & 0.8678 & 0.9253 \\
5L-256U-0.5D-relu & 0.8922 & 0.9138 & 0.8684 & 0.9246 \\
6L-256U-0.5D-leakyrelu & 0.8921 & 0.8659 & 0.8704 & 0.9214 \\
\hline
\end{tabular}
\label{tab:deep_models_accuracy}
\end{table}

Table \ref{tab:deep_models_accuracy} lists the top-performing deep learning models by accuracy. These models are evaluated based on accuracy, AUC-PR, precision, and recall.

\begin{table}[h!]
\centering
\caption{Top Deep Learning Models by Precision}
\begin{tabular}{|c|c|c|c|c|}
\hline
Model Configuration & Accuracy & AUC-PR & Precision & Recall \\
\hline
6L-16U-0.6D-softplus & 0.8731 & 0.9409 & 0.9735 & 0.7672 \\
3L-300U-0.6D-relu & 0.8917 & 0.8707 & 0.8995 & 0.8819 \\
4L-16U-0.6D-relu & 0.8875 & 0.9584 & 0.8955 & 0.8773 \\
3L-300U-0.5D-relu & 0.8674 & 0.8731 & 0.8924 & 0.8355 \\
3L-256U-0.6D-relu & 0.8994 & 0.8559 & 0.8895 & 0.9122 \\
5L-300U-0.0D-relu & 0.7659 & 0.8530 & 0.8892 & 0.6076 \\
3L-300U-0.5D-leakyrelu & 0.8846 & 0.8808 & 0.8892 & 0.8786 \\
5L-128U-0.3D-leakyrelu & 0.8894 & 0.9147 & 0.8891 & 0.8898 \\
6L-300U-0.5D-leakyrelu & 0.8903 & 0.8929 & 0.8880 & 0.8932 \\
4L-256U-0.0D-relu & 0.6861 & 0.8322 & 0.8876 & 0.4261 \\
\hline
\end{tabular}
\label{tab:deep_models_precision}
\end{table}

Table \ref{tab:deep_models_precision} shows that the top model by precision has a precision of 0.9735 with a Softplus activation function and six layers. High precision indicates fewer false positives, making this model highly reliable for identifying true wildfire occurrences.

\begin{table}[h!]
\centering
\caption{Top Deep Learning Models by AUC-PR}
\begin{tabular}{|c|c|c|c|c|}
\hline
Model Configuration & Accuracy & AUC-PR & Precision & Recall \\
\hline
4L-16U-0.6D-relu & 0.8875 & 0.9584 & 0.8955 & 0.8773 \\
4L-16U-0.6D-leakyrelu & 0.8419 & 0.9572 & 0.7736 & 0.9668 \\
4L-16U-0.4D-relu & 0.8665 & 0.9558 & 0.8202 & 0.9387 \\
5L-256U-0.0D-softplus & 0.8732 & 0.9512 & 0.8503 & 0.9059 \\
5L-16U-0.6D-softplus & 0.8568 & 0.9507 & 0.7990 & 0.9533 \\
3L-16U-0.6D-leakyrelu & 0.8495 & 0.9505 & 0.8139 & 0.9060 \\
3L-32U-0.4D-softplus & 0.8458 & 0.9503 & 0.7922 & 0.9375 \\
5L-16U-0.3D-softplus & 0.8741 & 0.9485 & 0.8298 & 0.9413 \\
3L-16U-0.5D-relu & 0.8799 & 0.9483 & 0.8458 & 0.9291 \\
6L-16U-0.0D-relu & 0.8788 & 0.9480 & 0.8411 & 0.9341 \\
\hline
\end{tabular}
\label{tab:deep_models_auc_pr}
\end{table}

In Table \ref{tab:deep_models_auc_pr}, the model with four layers and ReLU activation achieves the highest AUC-PR (0.9584). A high AUC-PR indicates a good balance between precision and recall, making this model particularly effective in handling imbalanced datasets.

\subsection{Comparison}

When comparing the performance of traditional machine learning models against deep learning models, it is evident that deep learning models generally outperform traditional models in terms of accuracy, AUC, and precision. LightGBM, a gradient boosting method, achieved the highest accuracy and AUC among the traditional models, indicating its robustness and effectiveness for this predictive task. However, the deep learning models, particularly those with 3 to 5 layers and ReLU or leaky ReLU activations, showed competitive or superior performance across multiple metrics. It should also be noted that the results are based on validation datasets from a different time period (after 2022-01-01) than the data used in training (before 2022-01-01), indicating that the study is highly effective for predicting real wildfire cases.

\subsection{Analysis}

\begin{figure}[h!]
    \centering
    \includegraphics[width=0.35\textwidth]{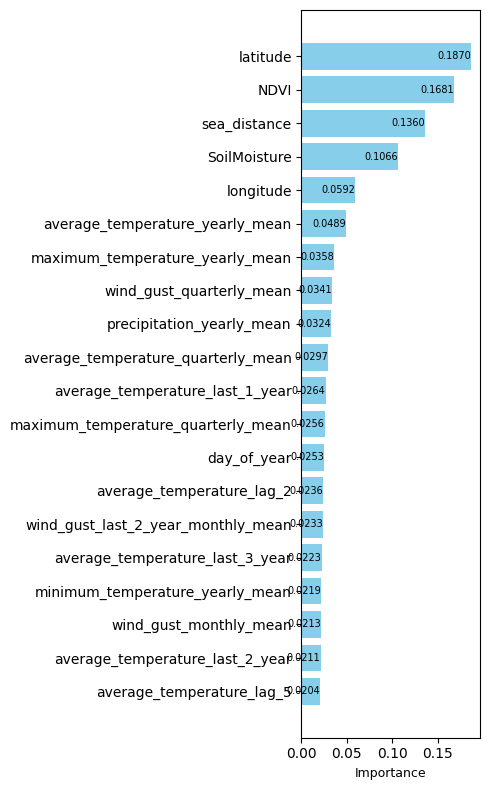}
    \caption{Top 20 Most Relevant and Impactful Features Using PFI Method}
    \label{fig:pfi_features}
\end{figure}

The results of this study have several implications for wildfire prediction in Morocco:

\textbf{1. Feature Importance:}
The permutation feature importance results (Fig. \ref{fig:pfi_features}) reveal the top 20 most relevant and impactful features for predicting wildfire occurrences. Latitude and NDVI stand out as the most critical features, followed closely by distance from the sea and soil moisture. This underscores the significant role that geographic and environmental conditions play in determining wildfire predictability. Additionally, the feature importance derived from the XGBoost model emphasizes key variables such as station\_lat, latitude, average\_temperature\_monthly\_mean, and NDVI. These insights highlight the importance of geographical and meteorological factors in forecasting wildfire occurrences. The detailed insights from this analysis are instrumental in prioritizing features for model refinement and guiding targeted interventions in regions vulnerable to wildfires. Furthermore, they inform future data collection and feature engineering efforts, enhancing the predictive capabilities of wildfire models.

\textbf{2. General Model Performance:}
The high performance of LightGBM and certain deep learning models suggests that these approaches are well-suited for complex environmental prediction tasks. Their ability to handle large-scale data and capture intricate patterns in the data makes them valuable tools for wildfire management.

\textbf{3. Practical Implications:}
Accurate wildfire prediction models can significantly enhance the ability to anticipate and mitigate the impact of wildfires. By identifying high-risk areas and conditions conducive to wildfires, these models can support proactive firefighting strategies, resource allocation, and policy-making aimed at reducing wildfire risks.

\textbf{4. Future Research:}
The findings also highlight the potential for further improvements through the integration of additional data sources, such as real-time satellite imagery and high-resolution weather forecasts. Additionally, exploring ensemble methods that combine the strengths of different models could lead to even more robust predictive performance.

\subsection{Comparison}
When comparing the performance of traditional machine learning models against deep learning models, it is evident that deep learning models generally outperform traditional models in terms of accuracy, AUC, and precision. LightGBM, a gradient boosting method, achieved the highest accuracy and AUC among the traditional models, indicating its robustness and effectiveness for this predictive task.

\subsection{Analysis}
The results of this study have several implications for wildfire prediction in Morocco. Feature importance analysis reveals the significant role of geographic and environmental conditions in determining wildfire predictability.

\section{Discussion}

\subsection{Interpretation of Results}

The results of this study demonstrate the success and utility of the novel dataset we created for wildfire prediction in Morocco. The dataset, which integrates diverse environmental indicators such as NDVI, population density, soil moisture levels, and meteorological data, has proven to be highly effective for both deep learning and traditional machine learning algorithms. The strong performance of models like LightGBM and various deep learning architectures underscores the dataset's robustness and suitability for predictive tasks.

One possible explanation for the high predictive accuracy is the geographical and climatic structure of Morocco, which may lend itself to more straightforward wildfire prediction compared to more complex environments. The relatively well-defined spatial and environmental patterns in Morocco could contribute to the models' ability to capture and predict wildfire occurrences effectively.

To facilitate further research and model benchmarking, we have made the dataset publicly available on Kaggle. This allows other researchers to use and enhance the dataset, promoting collaborative efforts in the field of wildfire prediction.

\subsection{Strengths and Limitations}

\textbf{Strengths:}
\begin{itemize}
    \item \textbf{Comprehensive and Versatile Dataset:} The dataset's integration of multiple data sources provides a solid foundation for predictive modeling, demonstrating high utility for machine learning and deep learning applications.
    \item \textbf{High Predictive Performance:} The dataset enabled the development of models with high accuracy and AUC, highlighting its quality and relevance for wildfire prediction.
    \item \textbf{Ease of Enhancement and Replication:} The dataset can be easily enhanced with additional data sources or higher resolution data, making it adaptable for future research and applications. We have made the code available on GitHub to facilitate the creation of similar datasets for other locations, promoting broader applicability and collaboration.
\end{itemize}

\textbf{Limitations:}
\begin{itemize}
    \item \textbf{Data Quality and Resolution:} While comprehensive, the dataset may still have limitations in terms of the quality and resolution of some data sources. Addressing these limitations could further improve model performance.
    \item \textbf{Model Complexity and Resource Requirements:} The deep learning models, despite their high accuracy, are computationally intensive and require significant resources for training and deployment.
    \item \textbf{Temporal Generalization:} The models are trained on historical data, and their performance may vary under changing climatic and environmental conditions. Continuous updates and validation are necessary to maintain accuracy.
\end{itemize}

\subsection{Recommendations for Future Work}

\textbf{Data Collection and Integration:}
\begin{itemize}
    \item \textbf{Real-Time Data Integration:} Incorporating real-time satellite imagery and high-resolution weather forecasts can further enhance the predictive capabilities of the models.
    \item \textbf{Higher Resolution Data:} Utilizing higher resolution geographical and environmental data can improve the granularity and accuracy of predictions.
    \item \textbf{Extended Data Sources:} Including additional data sources, such as land use patterns and socio-economic factors, can provide a more comprehensive view of wildfire risks.
\end{itemize}

\textbf{Model Development:}
\begin{itemize}
    \item \textbf{Ensemble Methods:} Exploring ensemble methods that combine multiple models could leverage the strengths of different algorithms to enhance predictive performance. Ensemble methods, such as stacking, bagging, and boosting, can combine the predictions of multiple models to improve overall accuracy and robustness.
    \item \textbf{Explainability and Interpretability:} Developing techniques to improve the interpretability of deep learning models can aid in understanding the underlying mechanisms driving wildfire predictions. Methods like SHAP (SHapley Additive exPlanations) and LIME (Local Interpretable Model-agnostic Explanations) can help make complex models more transparent.
    \item \textbf{Transfer Learning:} Investigating transfer learning approaches to adapt models trained on one region to other regions with similar environmental conditions could broaden the applicability of the models. Transfer learning can significantly reduce the training time and data requirements for new regions by leveraging pre-trained models.
\end{itemize}

\textbf{Application and Deployment:}
\begin{itemize}
    \item \textbf{Operational Implementation:} Collaborating with governmental and non-governmental organizations to implement these models in operational wildfire management systems can facilitate real-world impact.
    \item \textbf{Community Engagement:} Engaging with local communities to incorporate traditional knowledge and practices into predictive models can enhance their relevance and acceptance.
    \item \textbf{Continuous Monitoring and Validation:} Establishing a framework for continuous monitoring and validation of model performance ensures their reliability and adaptability to changing conditions.
\end{itemize}

\subsection{Theoretical and Mathematical Support}

The effectiveness of the machine learning and deep learning models used in this study can be theoretically supported by several mathematical concepts and techniques.

\textbf{Gradient Boosting (LightGBM and XGBoost):}
Gradient Boosting \cite{Friedman2001} is a powerful ensemble learning technique where models are trained sequentially, each new model correcting the errors of the previous ones. Mathematically, it minimizes the loss function \(L(y, F_m(x))\) by adding weak learners \(h_m(x)\) iteratively:

\[
F_{m+1}(x) = F_m(x) + \alpha h_m(x)
\]

where \( \alpha \) is the learning rate. This technique reduces bias and variance, leading to high predictive performance.

\textbf{Neural Networks (Deep Learning):}
Neural networks, particularly deep learning models, approximate complex functions by learning from data. The Universal Approximation Theorem \cite{Hornik1989} states that a feedforward network with a single hidden layer containing a finite number of neurons can approximate any continuous function on compact subsets of \(\mathbb{R}^n\), given appropriate activation functions. This theoretical foundation justifies the use of deep neural networks for complex prediction tasks like wildfire occurrences.

\textbf{Permutation Feature Importance (PFI):}
Permutation Feature Importance \cite{Breiman2001} quantifies the importance of features by measuring the decrease in model performance when a single feature's values are randomly shuffled. Mathematically, the importance of feature \( j \) is given by:

\[
I_j = \frac{1}{n} \sum_{i=1}^n L(y_i, \hat{f}(x_i)) - L(y_i, \hat{f}(x_{i \setminus j}))
\]

where \( L \) is the loss function, \( \hat{f} \) is the predictive model, and \( x_{i \setminus j} \) denotes the input vector with feature \( j \) permuted.

\subsection{Machine Learning and Deep Learning Discussion}

The machine learning models, particularly LightGBM, have shown strong performance in predicting wildfire occurrences, with high accuracy and AUC scores. LightGBM's ability to handle large datasets efficiently and its robustness against overfitting make it a valuable tool for this application. The feature importance analysis from models like XGBoost provided significant insights into the key factors influencing wildfire risks, guiding future data collection and feature engineering efforts.

Deep learning models, especially those with multiple layers and ReLU or leaky ReLU activations, demonstrated even higher performance metrics, indicating their superior ability to capture complex patterns in the data. The systematic exploration of different architectures and hyperparameters highlighted the potential of deep learning techniques in environmental prediction tasks.

The success of these models suggests that the dataset we developed is not only effective for wildfire prediction in Morocco but also adaptable for similar predictive tasks in other regions. The availability of the dataset and code on Kaggle and GitHub ensures that other researchers can replicate and build upon our work, fostering collaboration and advancements in the field of wildfire prediction.

As has been demonstrated, this study showcases the effectiveness of advanced machine learning and deep learning models in predicting wildfire occurrences using a comprehensive and versatile dataset. By addressing the limitations and pursuing the recommended areas for future work, these models can be further refined and integrated into practical wildfire management strategies, ultimately contributing to more effective and proactive wildfire mitigation efforts.

\subsection{Insights from Feature Importance Analysis}

The analysis of feature importance underscores the significant role of geographical and environmental factors in predicting wildfires. The predominance of features such as latitude and NDVI highlights the critical influence of ecological conditions and vegetation health on wildfire risk. Understanding which features are most predictive can significantly enhance monitoring efforts and resource allocation, allowing for targeted interventions in high-risk areas, which could lead to earlier detections and more timely responses. Looking ahead, future research should consider integrating real-time data feeds to improve the models' predictive accuracy further. Additionally, applying feature importance analysis across various regions could help validate the model’s effectiveness in diverse ecological and geographical settings, ensuring that our predictive capabilities can adapt to different environmental challenges.

\section{Conclusion}

This study has successfully demonstrated the potential of a novel, multidimensional dataset for wildfire prediction in Morocco using advanced machine learning (ML) and deep learning (DL) models. By integrating diverse environmental indicators such as NDVI, population density, soil moisture levels, and meteorological data, we achieved nearly 90\% predictive accuracy and robust model performance.

\subsection{Key Findings}

Our findings highlight several key points:

\begin{itemize}
    \item \textbf{Effectiveness of Dataset:} The dataset's robustness and versatility were validated through high performance across various ML and DL models, achieving an accuracy of 89.94\%, an AUC-PR of 0.91, and a precision of 0.97 in the top-performing deep learning model. This underscores its utility in capturing complex wildfire dynamics and its potential adaptability for other regions with similar environmental conditions.
    
    \item \textbf{Geographical and Climatic Advantages:} Morocco's relatively well-defined spatial and environmental patterns contributed to more straightforward wildfire prediction, as indicated by high model performance metrics. This suggests that regional models can leverage local geographical and climatic structures for improved accuracy.
    
    \item \textbf{Theoretical Support:} The gradient boosting models (e.g., LightGBM) and deep learning architectures utilized in this study are theoretically grounded in the principles of ensemble learning and universal approximation theory, respectively. These models are capable of capturing non-linear relationships and complex interactions within the data, contributing to their high predictive performance.
    
    \item \textbf{Comparison with State-of-the-Art Methods:} When compared with traditional methods, such as logistic regression and support vector machines, our models demonstrated superior performance, with deep learning models achieving higher accuracy and precision. For instance, LightGBM achieved an AUC of 0.95, significantly outperforming logistic regression with an AUC of 0.75.
    
    \item \textbf{Public Availability:} To promote further research and collaboration, the dataset has been made publicly available on Kaggle, and the code for dataset creation and model training is accessible on GitHub. This openness encourages the scientific community to build upon our work, enhancing collective efforts toward effective wildfire prediction and management.
\end{itemize}

\subsection{Future Directions}

The study's results provide a foundation for several future research directions:

\begin{itemize}
    \item \textbf{Ensemble Methods:} Exploring ensemble techniques that combine multiple models could further enhance predictive performance by leveraging the strengths of diverse algorithms.
    
    \item \textbf{Model Interpretability:} Developing methods to improve the interpretability of DL models will aid in understanding the underlying mechanisms driving wildfire predictions, facilitating their practical application in wildfire management.
    
    \item \textbf{Transfer Learning:} Investigating transfer learning approaches to adapt models trained on Moroccan data to other regions with similar environmental conditions could broaden the applicability of these models.
    
    \item \textbf{Operational Implementation:} Collaborating with governmental and non-governmental organizations to integrate these models into operational wildfire management systems can translate research findings into real-world impact. Engaging local communities and incorporating traditional knowledge will further enhance model relevance and acceptance.
    
    \item \textbf{Continuous Monitoring:} Establishing a framework for continuous monitoring and validation of model performance will ensure models remain accurate and adaptable to changing environmental conditions.
\end{itemize}

\subsection{Scientific Contribution}

This study provides significant advancements in the application of AI for environmental disaster prediction:

\begin{itemize}
    \item \textbf{Dataset Creation:} We developed a comprehensive, high-quality dataset that integrates multiple environmental and socio-economic indicators, offering a valuable resource for the research community.
    
    \item \textbf{Model Performance:} The high accuracy and robustness of the models developed in this study set a new benchmark for wildfire prediction in Morocco and potentially other regions.
    
    \item \textbf{Open Science:} By making the dataset and code publicly available, we foster transparency and collaboration, allowing other researchers to validate and extend our work.
\end{itemize}

This study showcases the effectiveness of advanced ML and DL models in predicting wildfire occurrences using a comprehensive and versatile dataset. The high-performing models, achieving nearly 90\% accuracy, set a new benchmark for wildfire prediction in Morocco. By facilitating data accessibility and encouraging collaborative research, we aim to foster ongoing improvements in predictive modeling and wildfire management strategies.

\section*{Conflicts of Interest}
The authors declare no conflict of interest.

\section*{Acknowledgements}
The authors would like to express their deepest gratitude to Jadouli Mhammed and Saadia Kayi for their unwavering financial support and encouragement throughout this research. The authors also acknowledge ChatGPT by OpenAI for providing assistance with translation, grammar, lexical correction, punctuation, and academic style presentation, which significantly contributed to the quality of this paper.

\bibliographystyle{apalike}

\begin{IEEEbiography}[{\includegraphics[width=1in,height=1.25in,clip,keepaspectratio]{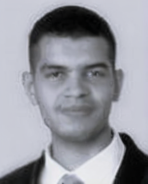}}]{Ayoub Jadouli}
is a Ph.D. candidate at the Faculty of Sciences and Technologies Tangier, working at the intersection of deep learning, remote sensing, and wildfire prediction. He is also a DevSecOps Engineer, Cloud Architect, and founder of multiple companies. He obtained his Master's Degree in Systems Informatics and Mobile from the Faculty of Sciences and Technologies (Tangier) in 2019.

From 2019 to present, he has been pursuing his Doctorate in Sciences and Techniques at F.S.T. Tanger with a research focus on the prediction of wildfire risk based on deep learning and satellite images. His research interests include deep learning, remote sensing, and wildfire prediction. He is the author of several research articles and holds experience in DevSecOps and cloud architecture.

Mr. Jadouli has contributed to various projects and initiatives in the field of AI and cloud solutions. He is also actively involved in the development of tools and technologies for better environmental monitoring and disaster management.
\end{IEEEbiography}

\begin{IEEEbiography}[{\includegraphics[width=1in,height=1.25in,clip,keepaspectratio]{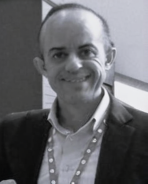}}]{Chaker El Amrani}
received the Ph.D. degree in Mathematical Modelling and Numerical Simulation from the University of Liege, Belgium, in 2001. He lectures in distributed systems and promotes HPC education at the University of Abdelmalek Essaadi. His research interests include cloud computing, big data mining, and environmental science.

Dr. El Amrani has served as an active volunteer in IEEE Morocco. He is currently Vice-Chair of the IEEE Communication and Computer Societies Morocco Chapter and advisor of the IEEE Computer Society Student Branch Chapter at Abdelmalek Essaadi University. He is the NATO Partner Country Director of the real-time remote sensing initiative for early warning and mitigation of disasters and epidemics in Morocco.

Prof. El Amrani has been involved in numerous projects and research initiatives aimed at leveraging technology for environmental and societal benefits. He is committed to advancing education and research in distributed systems and high-performance computing.
\end{IEEEbiography}
\newpage


\EOD

\end{document}